\documentclass[letterpaper,11pt]{article}
\usepackage{coda_preprint}

\usepackage[T1]{fontenc}
\usepackage{amsmath,amssymb,graphicx,booktabs,float}
\usepackage[stretch=30,shrink=30]{microtype}
\usepackage{cite}
\usepackage[hyphens]{url}
\usepackage{xcolor}

\usepackage[hidelinks,hyperfootnotes=false,hypertexnames=false]{hyperref}
\newcommand{\paperpara}[1]{\par\smallskip\noindent\textbf{#1.}\enspace\ignorespaces}

\title{Fewer Steps, Better Actions:\\ Rethinking Flow-Matching Inference for VLA Policies}
\author{
Zhipeng Tang$^{\clubsuit\diamond}$\quad Xinda Chen$^{\clubsuit\diamond}$\quad
Weining Rao$^{\clubsuit\diamond}$\quad Xiao Li$^{\clubsuit\diamond}$\\[5pt]
Wenting Tan$^{\clubsuit}$\quad Yuning Wang$^{\triangle}$\quad
Xiao Shi$^{\clubsuit}$\thanks{Corresponding author.}\quad Xiaofang Zhao$^{\clubsuit}$\\[10pt]
\small$^{\clubsuit}$Institute of Computing Technology, Chinese Academy of Sciences, Beijing, China\\
\small$^{\diamond}$University of Chinese Academy of Sciences, Beijing, China\\
\small$^{\triangle}$Zhongke Haichuan Intelligent\\[6pt]
\small\texttt{\{tangzhipeng23s, chenxinda24z, shixiao\}@ict.ac.cn}
}
\date{}
\hypersetup{pdftitle={Fewer Steps, Better Actions: Rethinking Flow-Matching Inference for VLA Policies},pdfauthor={Zhipeng Tang, Xinda Chen, Weining Rao, Xiao Li, Wenting Tan, Yuning Wang, Xiao Shi, Xiaofang Zhao}}
\begin{document}
\maketitle

\begingroup
\renewcommand{\thefootnote}{}
\footnotetext{This work has been submitted to the IEEE for possible publication. Copyright may be transferred without notice, after which this version may no longer be accessible.}
\endgroup
\begin{abstract}
Vision-language-action (VLA) policies based on flow matching generate action chunks through repeated evaluations of an action expert. Increasing the number of integration steps raises inference cost, but does not necessarily improve closed-loop success. We propose Coda, which reallocates part of this integration budget to a single learned endpoint correction. A frozen policy first completes a few-step noise-to-action trajectory; a lightweight Transformer then predicts a demonstration-supervised residual using the candidate action, source noise, and shared observation-prefix cache. Only the corrector is trained. On 50 RoboTwin Easy tasks, five-step Coda improves success from 71.64\% to 74.68\% over the matched five-step baseline, while reducing forward latency by 30.2\% relative to the default ten-step policy. A two-step configuration achieves 71.88\% success with a 2.12$\times$ speedup. An independent 13-task control shows a 5.69-percentage-point gain at nearly equal latency, supporting correction as an effective alternative to additional integration. The same design also improves frozen official SmolVLA, raising two-step success from 60.8\% to 69.4\%. These results show that endpoint correction improves the quality--latency trade-off of frozen flow-matching policies.

\end{abstract}

\section{Introduction}
\label{sec:intro}
Vision-language-action (VLA) policies map images, language instructions, and robot states to actions, providing a unified interface for multitask manipulation~\cite{brohan2022rt1,zitkovich2023rt2,kim2024openvla,openx2024}. Flow-matching policies such as $\pi_0$, $\pi_{0.5}$, and SmolVLA first encode the observation condition and then repeatedly invoke an action expert (AE) to integrate noise into an action chunk~\cite{black2024pi0,intelligence2025pi05,shukor2025smolvla}. Each step depends on the output of the previous one, so action-chunk generation requires multiple serial AE forwards. The number of function evaluations (NFE) directly drives the serial AE cost~\cite{lipman2023flow,liu2023rectified}. In our A100 profiling, the AE stage accounts for about three quarters of the default ten-step forward pass. Reducing this repeated computation is therefore a major opportunity for lowering model-forward latency.

Do more integration steps lead to more successful actions? In our $50$-task evaluation, increasing the number of integration steps from one to five raises success from $59.76\%$ to $71.64\%$, whereas further increasing it to ten steps yields $71.52\%$. This plateau motivates our central question: \emph{if the benefit of continued serial integration is limited, can part of its time and compute budget be reassigned to a single learned correction?}
\begin{figure}[htbp]
\centering
\includegraphics[width=.86\linewidth]{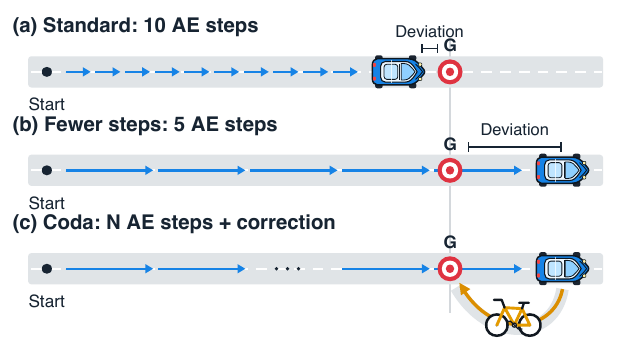}
\caption{Compute allocation. Ten-step and few-step generation both cover the full noise-to-action interval. Coda reallocates part of the serial integration budget to one learned endpoint correction; the target and offsets are schematic.}
\label{fig:method}
\end{figure}

Few-step generation and correction play complementary roles. Reducing ten Euler steps to five increases the step size from $1/10$ to $1/5$, advancing farther along the 
velocity field at each step. Both settings cover the complete noise-to-action interval. This coarser integration saves time but may change the final action. Coda uses the few-step candidate and spends part of the saved budget on one learned endpoint correction. Fig.~\ref{fig:method} illustrates this change in computation using a car and a bicycle.

A preliminary endpoint intervention further motivates this design. On \textrm{hanging\_mug}, mixing an expert-reference action with source noise at $t=0.1$ and applying one AE step succeeds in 9/10 trials, compared with 5/50 for the five-step baseline (Fig.~\ref{fig:diagnostics}a). This privileged probe suggests that endpoint information can improve the final action. Coda learns such a correction from demonstrations, so deployment requires only the policy's own candidate and observation.

We propose Coda, which separates action generation from endpoint correction. Named after the concluding passage of a musical composition, Coda adds a single learned correction after the complete few-step generation trajectory. The original policy uses $N$ integration steps to generate a candidate $a_\theta$, and an independent Transformer predicts a residual $\widehat{\Delta a}$. The executed action is
\begin{equation}
a_{\mathrm{exec}}=a_\theta+\widehat{\Delta a}.
\label{eq:exec}
\end{equation}
The corrector is trained to predict the difference between the demonstrated action and the candidate produced by the frozen generator. During inference, it reuses the key-value (KV) cache from the observation encoding, which limits the additional computation~\cite{pope2023efficiently}. Only the corrector is trained; the generator remains frozen.

The corrector conditions on the candidate action, observation context, and source noise. Its single forward pass is separate from integration, so the generation step count can be chosen to meet the available inference budget.

We evaluate both closed-loop success and measured inference latency. On the 50-task main panel, five-step Coda gains 3.04 percentage points over five-step generation and is 1.43$\times$ faster than the default ten-step policy. In an independent 13-task panel, five-step Coda outperforms six-step generation by 5.69 points at nearly identical latency. A second implementation on frozen official SmolVLA tests whether this design extends beyond $\pi_{0.5}$.

Our contributions are:
\begin{enumerate}
\item \textbf{A compute-allocation strategy for few-step generation.}
We show that higher NFE substantially increases serial AE computation but provides limited additional closed-loop benefit in the evaluated setting. Based on this observation, Coda replaces part of the integration budget with a single endpoint correction.

\item \textbf{A configurable corrector for frozen flow-matching policies.}
Coda keeps the original policy frozen, reuses its observation-prefix representation, and learns the residual between the generated candidate and the demonstrated action. A single corrector can be applied across test-time NFE values, and the same design is implemented on both $\pi_{0.5}$ and SmolVLA.

\item \textbf{An evaluation of both action quality and inference cost.}
On the $50$-task RoboTwin suite, Coda improves mean success over matched-NFE baselines, while selected few-step configurations reduce latency relative to the default ten-step policy. Cross-NFE experiments, transfer to SmolVLA, and comparisons with Mix-AE, continued AE training, and an MLP corrector characterize how endpoint correction uses the available inference budget.
\end{enumerate}
\section{Related Work}
\paragraph{Generative action policies and deployment latency}
Diffusion models, score-based SDEs, and flow matching model continuous action distributions through iterative transformations~\cite{ho2020ddpm,song2021sde,lipman2023flow,liu2023rectified,tong2024cfm}. Diffusion Policy, 3D Diffusion Policy, ACT, Octo, and RDT-1B combine generative action modeling with action-chunk execution for visuomotor control, bimanual manipulation, and multitask policies~\cite{chi2023diffusion,ze2024dp3,zhao2023act,octo2024,liu2024rdt}. RT-1, RT-2, and OpenVLA instead use autoregressive or discrete-action VLA interfaces, while Open X-Embodiment provides cross-embodiment training data~\cite{brohan2022rt1,zitkovich2023rt2,kim2024openvla,openx2024}. $\pi_0$, $\pi_{0.5}$, and SmolVLA combine vision-language conditioning with flow-matching action experts and generate action chunks through repeated action-expert (AE) evaluations; $\pi_{0.5}$ uses a PaliGemma-based vision-language prefix~\cite{black2024pi0,intelligence2025pi05,shukor2025smolvla,beyer2024paligemma}. Their planning latency therefore includes both observation-prefix encoding and NFE-proportional AE computation. Real-Time Execution of Action Chunking Flow Policies mitigates chunk-arrival delay through asynchronous generation, overlapping execution, and chunk alignment, while retaining the underlying iterative action generator~\cite{black2025rtc}. Coda addresses a different layer: how to allocate inference compute between candidate generation and endpoint correction.

\paragraph{Few-step generation and sampling acceleration}
For diffusion sampling, DDIM uses a non-Markovian sampling process and DPM-Solver uses a dedicated high-order ODE solver to reduce sampling steps~\cite{song2021ddim,lu2022dpm}; progressive distillation, consistency models, latent consistency models, and shortcut models support few- or one-step generation through distillation or new training objectives~\cite{salimans2022progressive,song2023consistency,luo2023lcm,frans2025shortcut}. In visuomotor control, Consistency Policy applies consistency distillation to few-step inference~\cite{prasad2024consistency}. In flow matching and robotic action generation, Rectified Flow improves few-step transport by straightening transport trajectories, while L1 Sample Flow combines an integration step with direct sample prediction under an $L_1$ objective~\cite{liu2023rectified,song2025l1flow}. These approaches primarily improve the solver, alter the generation trajectory, or retrain a few-step generation objective.

\paragraph{Generation sources, residual policies, and post-generation correction}
Action-to-action flow matching encodes proprioceptive and action history into the generation source, changing the action distribution from its starting point~\cite{jia2026a2a}; ResVLA anchors generation at a predicted low-frequency intent and models the remaining high-frequency dynamics with a residual diffusion bridge~\cite{zhong2026resvla}. Classical residual policy learning adds a learned action correction outside a fixed controller~\cite{silver2018rpl,johannink2019rrl}, a form also related to deep residual connections~\cite{he2016resnet}. 

\paragraph{Position of this work}
Coda applies supervised residual learning to a frozen flow-matching VLA at a specific interface: the endpoint of a complete few-step trajectory. It reuses the prefix KV cache, conditions on the generated candidate, and decouples the integration budget from the correction forward. Our focus is the resulting compute allocation: whether a learned endpoint update delivers better closed-loop quality per unit inference time than additional calls to the original action expert.

\begin{figure}[htbp]
\centering
\includegraphics[width=1.0\linewidth]{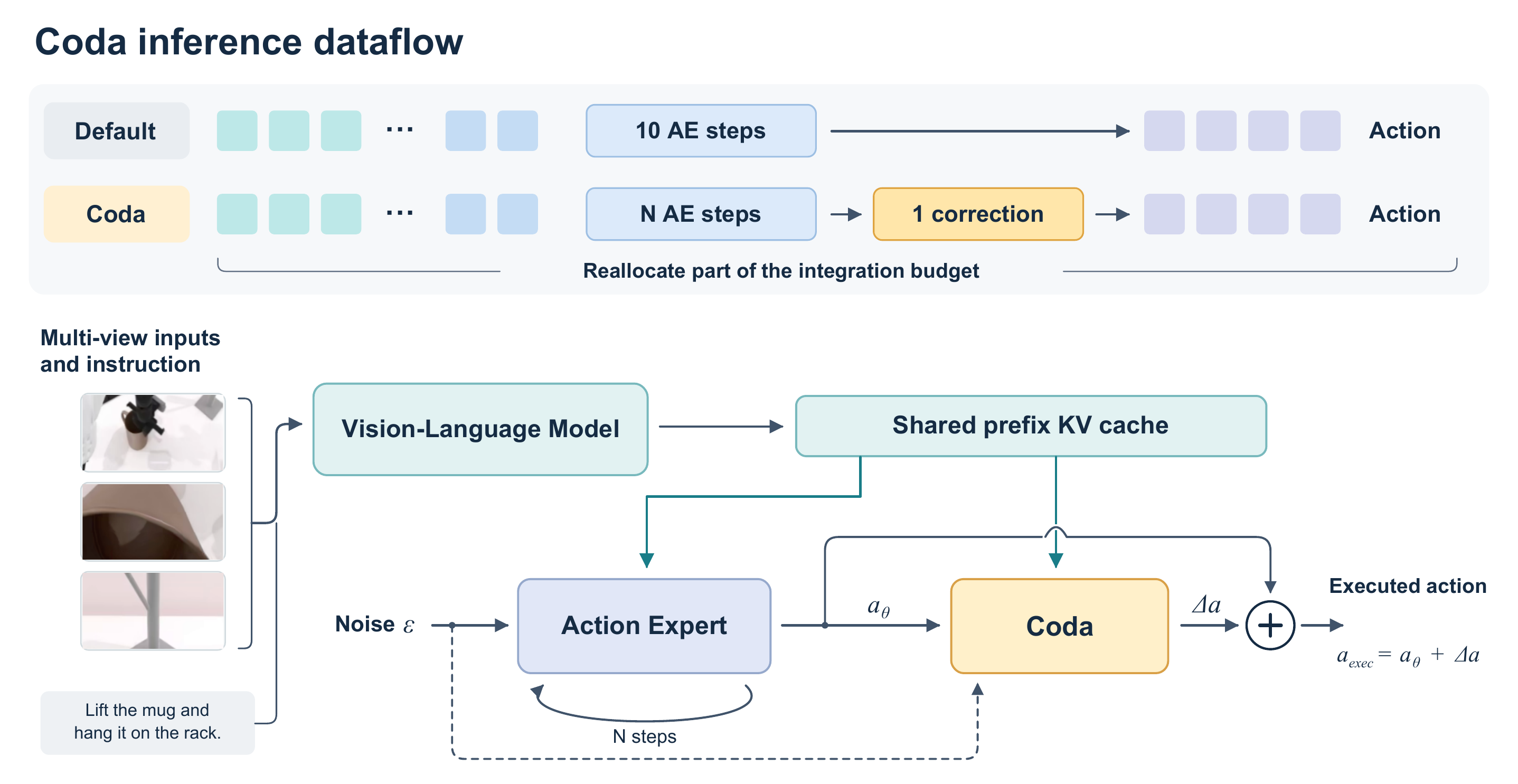}
\caption{Coda inference. A frozen VLM encodes the three camera views and instruction once; in $\pi_{0.5}$, discretized robot state is also included in the text prefix. Both modules reuse the prefix KV cache. The frozen AE completes $N$ Euler steps, and Coda uses the candidate and the same source noise to predict a residual in one forward. The residual is added to the candidate before execution. A line bridge denotes a crossing without a connection. Only Coda is trainable.}
\label{fig:architecture}
\end{figure}

\section{Method}
\label{sec:method}
\subsection{Few-step generation on frozen \texorpdfstring{$\pi_{0.5}$}{pi0.5}}
Let $o$ denote the observation and instruction, and let $a^\star\in\mathbb{R}^{H\times D}$ be an expert action chunk, where $H$ is the prediction horizon and $D$ is the action dimension. The source noise is $\varepsilon\sim\mathcal{N}(0,I)$. We use the convention that $t=1$ is noise and $t=0$ is action. The flow-matching training interpolation is
\begin{equation}
x_t=t\varepsilon+(1-t)a^\star,
\label{eq:interp}
\end{equation}
with target velocity $\varepsilon-a^\star$~\cite{lipman2023flow,liu2023rectified,tong2024cfm}. Let $z_o=f_{\mathrm{VLM}}(o)$ be the frozen observation representation. Inference uses $N$ Euler steps:
\begin{equation}
x^{[i+1]}=x^{[i]}-\frac{1}{N}v_\theta(x^{[i]},1-i/N,z_o),
\label{eq:euler}
\end{equation}
where $i=0,\ldots,N-1$, $x^{[0]}=\varepsilon$, and the final candidate is $a_\theta=x^{[N]}$. Reducing $N$ increases the integration step size. Coda and its matched-NFE baseline use the same integration process. Fig.~\ref{fig:architecture} shows this $N$-step path to $a_\theta$ and the residual suffix attached after the candidate.

\subsection{Candidate-conditioned residual learning}
The residual module conditions on the frozen observation prefix, the candidate action, and its source noise:
\begin{equation}
\widehat{\Delta a}=r_\phi(z_o,a_\theta,\varepsilon,t_{\mathrm{res}}).
\end{equation}
On $\pi_{0.5}$, the candidate action and noise are projected separately and added to form the input to a Transformer~\cite{vaswani2017attention} suffix. As shown in Fig.~\ref{fig:architecture}, the residual suffix reads the VLM prefix KV cache, while the prefix does not read the residual suffix. We set $t_{\mathrm{res}}=0.1$ as fixed time conditioning through AdaRMS modulation, matching the near-endpoint time used in GT-mix. This fixed conditioning value is used by the corrector after generation reaches $t=0$. SmolVLA omits this time conditioning. The residual Transformer suffix is initialized from the corresponding frozen AE weights, and its output layer is zero-initialized so that the initial correction is zero and training starts from $a_{\mathrm{exec}}=a_\theta$.

Training updates only residual parameters $\phi$; the VLM and AE remain frozen throughout. With $\operatorname{sg}$ denoting stop-gradient, the supervision target and loss are
\begin{align}
\Delta a^\star&=a^\star-\operatorname{sg}(a_\theta),\\
\mathcal{L}_{\mathrm{res}}&=\left\|r_\phi(z_o,\operatorname{sg}(a_\theta),\varepsilon,t_{\mathrm{res}})-\Delta a^\star\right\|_1.
\label{eq:l1}
\end{align}
This objective learns the total displacement from the generated candidate to the demonstrated action, including both policy bias and the effect of coarse integration. The $\pi_{0.5}$ residual is trained on candidates with $N_{\mathrm{train}}=5$ and the same weights are reused at different test NFE values.

For SmolVLA, we retain the same candidate-plus-residual form and train an independent corrector on the frozen SmolVLA policy. This implementation uses $N_{\mathrm{train}}=10$, omits the time conditioning, and applies causal attention to residual tokens. Residual weights are not shared across policies.

\subsection{Direct execution and inference cost}
At inference, Coda outputs the corrected action according to~\eqref{eq:exec}, executes the first $K$ actions of the predicted chunk, and then replans from a new observation. The executed horizon is $K$; evaluation uses $K=10$ for $\pi_{0.5}$ and $K=50$ for SmolVLA (Table~\ref{tab:env}). Each replan encodes the observation prefix once and links generation and correction with the same source noise. Expert actions are used only as training supervision.

If prefix, one AE forward, and one residual-forward time are $T_{\mathrm{prefix}}$, $T_{\mathrm{AE}}$, and $T_{\mathrm{res}}$, respectively, then
\begin{align}
T_{\pi_{0.5}\text{-FT}}(N)&\approx T_{\mathrm{prefix}}+NT_{\mathrm{AE}},\\
T_{\mathrm{Coda}}(N)&\approx T_{\mathrm{prefix}}+NT_{\mathrm{AE}}+T_{\mathrm{res}}.
\end{align}
Thus the net saving from ten-step generation to five-step Coda is approximately $5T_{\mathrm{AE}}-T_{\mathrm{res}}$. The speedup comes from reducing repeated AE calls; matched-NFE quality improvement incurs one correction forward. Below, NFE counts AE calls only, and Coda's single residual forward is reported separately.

\section{Experiments}
\subsection{Experimental setup}
\paperpara{Main evaluation}
We evaluate on 50 RoboTwin bimanual manipulation tasks~\cite{mu2025robotwin,chen2025robotwin2}. The base $\pi_{0.5}$ policy is fine-tuned for 60k steps on \textrm{demo\_clean\_50} and then frozen ($\pi_{0.5}$-FT). Coda is trained for 3k steps on this checkpoint. Global batch sizes are 48 and 960, respectively, giving 2.88M presented samples for each stage. Peak AdamW learning rates are $2.5\times10^{-5}$ and $5\times10^{-5}$, with 3000 and 150 warmup steps and cosine decay. Execution uses the first $K=10$ actions before replanning.

For each NFE in $\{1,2,3,4,5,10\}$, both methods complete 50 valid episodes on each task (2500 episodes per method). Additional stable seeds fill the six tasks whose default initialization pools contain fewer than 50 valid episodes. Success is the equally weighted task mean. We obtain the 95\% interval for the five-step gain by resampling paired task differences 100000 times~\cite{efron1979bootstrap}. The source-noise stream is determined by the environment seed and replan index. Complete execution and environment settings appear in the appendix.

\paperpara{Additional evaluations}
Official SmolVLA remains frozen, and a separate corrector is trained for 18716 steps on its official 10-task set, using batch size 256 and peak learning rate $10^{-4}$. Evaluation uses 50 episodes per task and execution horizon $K=50$. The no-$\varepsilon$ input variant uses checkpoint 18000. An independent 13-task panel contains 621 valid episodes per method and compares Transformer correction, MLP correction, and additional AE integration at nearby inference budgets.

\paperpara{Latency}
The main operating points use an A100 80GB PCIe, batch size one, SDPA, and no compilation. We report the median of 40 CUDA-synchronized model forwards after 10 warmups, including prefix encoding, all AE calls, and correction. These timings exclude environment, communication, and robot execution. All $\pi_{0.5}$ main points come from the same measured NFE sweep. The independent 13-task panel uses an A100 40GB and its own observation cache and timing protocol.

\begin{table}[htbp]
\caption{Success on 50 Easy tasks (\%). Each method has 50 episodes per task at every NFE. Bold indicates the higher mean at each NFE.}
\label{tab:main}
\centering\small\setlength{\tabcolsep}{3pt}
\begin{tabular}{lrrrrrr}\toprule
\textbf{AE-NFE} & \textbf{1} & \textbf{2} & \textbf{3} & \textbf{4} & \textbf{5} & \textbf{10}\\\midrule
$\pi_{0.5}$-FT & 59.76 & 67.40 & 71.28 & 69.96 & 71.64 & 71.52\\
Coda & \textbf{69.80} & \textbf{71.88} & \textbf{73.64} & \textbf{73.12} & \textbf{74.68} & \textbf{72.60}\\\bottomrule
\end{tabular}
\end{table}

\subsection{More integration has diminishing closed-loop returns}
\begin{figure}[htbp]
\centering
\includegraphics[width=.76\linewidth]{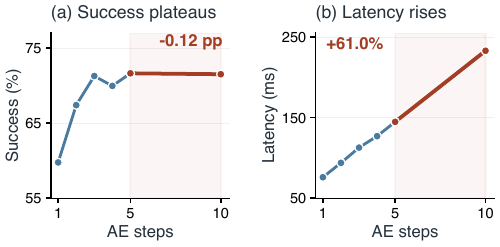}
\caption{Diminishing returns of frozen $\pi_{0.5}$ on the 50-task panel. (a) Success versus AE steps. (b) Measured forward latency versus AE steps. The highlighted 5-to-10-step segment changes success by $-0.12$ percentage points while increasing latency by 61.0\%.}
\label{fig:nfe-sweep}
\end{figure}

The frozen baseline improves from 59.76\% at one step to 71.64\% at five steps, but reaches 71.52\% at ten steps (Table~\ref{tab:main}). Over the latter interval, measured latency rises from 144.7 to 233.0\,ms, a 61.0\% increase (Fig.~\ref{fig:nfe-sweep}). Thus, increasing integration resolution beyond five steps adds substantial serial computation without improving mean success. This observation motivates using part of that computation for an endpoint update learned directly from demonstration actions.

Five-step Coda reaches 74.68\%, a 3.04-percentage-point improvement over five-step generation, with a task-bootstrap interval of [1.04, 5.08]. Across tasks, 27 improve, 10 tie, and 13 decline. The same corrector yields positive mean gains at every tested NFE, from 10.04 points at one step to 1.08 points at ten steps. Five steps give the highest success among the tested Coda configurations.

\subsection{Correction is a better use of comparable inference compute}
\begin{figure}[htbp]
\centering
\includegraphics[width=.76\linewidth]{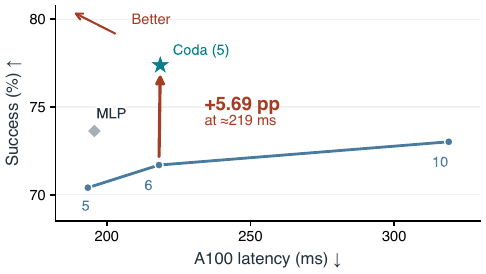}
\caption{Where to spend the inference budget. The baseline line connects measured 5-, 6-, and 10-step policies. At approximately 219\,ms, Coda improves success by 5.69 points over six-step generation. Upper-left is better; this panel uses the independent 13-task protocol in Table~\ref{tab:budget}.}
\label{fig:budget}
\end{figure}

The independent 13-task panel compares correction with additional integration at nearly equal latency. Six-step $\pi_{0.5}$-FT reaches 71.69\% at 218.14\,ms. Five-step Coda reaches 77.38\% at 218.59\,ms: a 5.69-point improvement at nearly equal latency (Fig.~\ref{fig:budget}). The paired 95\% interval is [2.62, 8.77] points when resampling episodes within fixed tasks and [1.54, 10.15] under two-stage task-and-episode resampling. This comparison supports allocating the next unit of computation to endpoint correction rather than another AE step.

The same Coda point also exceeds ten-step generation by 4.37 points while using 31.5\% less forward time. An MLP corrector reaches 73.63\% at 195.63\,ms, providing a cheaper intermediate option. The MLP and Transformer correctors offer different quality--latency choices at the same integration step count.

\begin{table}[htbp]
\caption{Independent 13-task compute-allocation panel. Each method has 621 valid episodes; timing uses an A100 40GB.}
\label{tab:budget}\centering\small
\begin{tabular}{lrrr}\toprule
\textbf{Method} & \textbf{AE-NFE} & \textbf{Success (\%)} & \textbf{Latency (ms)}\\\midrule
$\pi_{0.5}$-FT & 5 & 70.40 & 193.39\\
$\pi_{0.5}$-FT & 6 & 71.69 & 218.14\\
$\pi_{0.5}$-FT & 10 & 73.02 & 318.98\\
Coda-MLP & 5 & 73.63 & 195.63\\
Coda-Transformer & 5 & \textbf{77.38} & 218.59\\\bottomrule
\end{tabular}\end{table}

\subsection{Coda improves the deployment trade-off}
\begin{figure}[htbp]
\centering
\includegraphics[width=.76\linewidth]{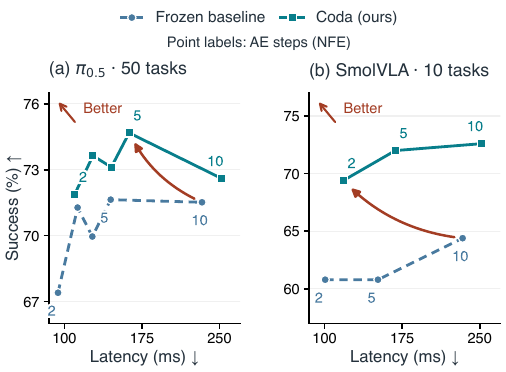}
\caption{Deployment landscapes on A100 80GB: (a) $\pi_{0.5}$, 50 tasks; (b) frozen official SmolVLA, 10 tasks. Upper-left is better. Point labels (2, 5, 10) indicate the number of AE integration steps (NFE); Coda adds one correction forward at each setting. Lines connect measured points at NFE$\geq2$, and arrows highlight improvements over the default ten-step policy. One-step success is retained in Tables~\ref{tab:main} and~\ref{tab:smol}. The panels use separate host policies and evaluation protocols.}
\label{fig:latencyquality}
\end{figure}

Fig.~\ref{fig:latencyquality} maps the main evaluation to measured success--latency operating points.  Each line connects measured configurations of one method; step counts identify the deployment budget.

For $\pi_{0.5}$, five-step Coda reaches 74.68\% at 162.6\,ms, compared with 71.52\% at 233.0\,ms for the default ten-step policy. It therefore improves success by 3.16 points while reducing latency by 30.2\% (1.43$\times$ speedup). Two-step Coda offers a lower-cost point: 71.88\% at 109.9\,ms, a 2.12$\times$ speedup over the default. It also exceeds the success of three-step generation (71.28\% at 112.7\,ms) at slightly lower latency.

Table~\ref{tab:latency} separates the cost of generation and correction. At five steps, Coda adds 17.9\,ms to the 144.7\,ms baseline. Reducing AE calls from ten to five more than covers that addition. Prefix encoding takes approximately 56\,ms and correction approximately 18\,ms, so these two costs become a larger fraction of the total at low NFE. The resulting operating points support choosing a two-step configuration for lower latency or a five-step configuration for higher success, using the same trained corrector.
\begin{table}[htbp]
\caption{Measured median forward latency (ms). P/A/R denote prefix, AE, and residual stage times. Speedup is relative to the same-GPU ten-step baseline. Bold highlights Coda deployment points.}
\label{tab:latency}\centering\small\setlength{\tabcolsep}{2.5pt}
\begin{tabular}{llrrrrrr}\toprule
\textbf{GPU} & \textbf{Method} & \textbf{N} & \textbf{P} & \textbf{A} & \textbf{R} & \textbf{Total} & \textbf{Speedup}\\\midrule
A100 & $\pi_{0.5}$-FT & 10 & 55.8 & 175.6 & 0.0 & 233.0 & $1.00\times$\\
 & $\pi_{0.5}$-FT & 5 & 55.8 & 88.0 & 0.0 & 144.7 & $1.61\times$\\
 & \textbf{Coda} & 5 & 55.7 & 88.0 & 17.8 & \textbf{162.6} & {\bfseries\boldmath $1.43\times$}\\
 & $\pi_{0.5}$-FT & 3 & 57.1 & 54.3 & 0.0 & 112.7 & $2.07\times$\\
 & \textbf{Coda} & 2 & 55.9 & 35.3 & 17.9 & \textbf{109.9} & {\bfseries\boldmath $2.12\times$}\\
\midrule
4090 & $\pi_{0.5}$-FT & 10 & 62.7 & 182.1 & 0.0 & 246.5 & $1.00\times$\\
 & \textbf{Coda} & 5 & 61.7 & 88.6 & 17.8 & \textbf{169.2} & {\bfseries\boldmath $1.46\times$}\\
 & \textbf{Coda} & 2 & 61.9 & 35.6 & 17.9 & \textbf{116.2} & {\bfseries\boldmath $2.12\times$}\\
\bottomrule\end{tabular}
\par\smallskip\small Stage times and total latency are measured separately; medians are not additive.
\end{table}

\subsection{The design transfers to frozen SmolVLA}
On official SmolVLA, Coda raises success from 60.8\% to 69.4\% at two steps, from 60.8\% to 72.0\% at five steps, and from 64.4\% to 72.6\% at ten steps (Table~\ref{tab:smol}). The two-step corrected policy runs at 118.1\,ms, compared with 233.5\,ms for ten-step official generation: a 1.98$\times$ speedup with a 5.0-point gain. Fig.~\ref{fig:latencyquality}b shows the same quality--latency improvement on SmolVLA.

\begin{table}[htbp]
\caption{SmolVLA success on 10 Easy tasks (\%). Each configuration completes 500 episodes. AE-cont and Coda ($+\varepsilon$) use checkpoint 18716; the no-$\varepsilon$ variant uses checkpoint 18000.}
\label{tab:smol}\centering\small\setlength{\tabcolsep}{4pt}
\begin{tabular}{lrrrr}\toprule
\textbf{AE-NFE} & \textbf{1} & \textbf{2} & \textbf{5} & \textbf{10}\\\midrule
Official frozen & 35.0 & 60.8 & 60.8 & 64.4\\
AE-cont & 48.2 & 64.6 & 70.8 & 69.2\\
Coda ($+\varepsilon$) & 55.8 & 69.4 & 72.0 & 72.6\\
Coda (no $\varepsilon$) & \textbf{58.2} & \textbf{71.4} & \textbf{75.2} & \textbf{74.8}\\\bottomrule
\end{tabular}\end{table}

The SmolVLA corrector is trained on ten-step candidates and directly reused at one, two, and five steps. The $\pi_{0.5}$ corrector is trained at five steps and reused over the full tested range. This cross-NFE reuse allows deployment budgets to change without training a new corrector for each setting.

AE continuation is an alternative use of additional training compute. Under the matched SmolVLA training budget, continuation reaches 64.6/70.8/69.2\% at two/five/ten steps, below the corresponding Coda results. The independently trained no-$\varepsilon$ configuration reaches 58.2/71.4/75.2/74.8\%, indicating that source-noise input is unnecessary for this SmolVLA configuration. The deployment plot uses the noise-conditioned model consistently across all budgets.

\subsection{Direct execution preserves the learned correction}
We compare direct execution with Mix-AE, which sends the corrected action through one additional frozen AE update. It first constructs
\begin{equation}
\widetilde{x}_{0.1}=0.1\varepsilon+0.9(a_\theta+\widehat{\Delta a}),
\label{eq:mix}
\end{equation}
then integrates from $t=0.1$ to zero using the original velocity field. The same source noise is reused in this remapping.

\begin{figure}[htbp]
\centering
\includegraphics[width=.82\linewidth]{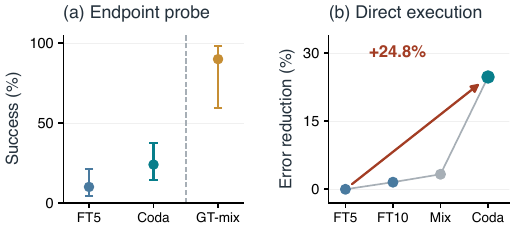}
\caption{Endpoint diagnostics. (a) Hanging-mug success with 95\% Wilson intervals; the dashed divider separates the privileged GT-mix probe from deployable policies. (b) Reduction in mean open-loop action error relative to FT5; Mix denotes Mix-AE. Direct Coda execution preserves the largest reduction.}
\label{fig:diagnostics}
\end{figure}

On 128 held-out mix50 demonstration frames, mean absolute error over 12 arm joints is $5.73\times10^{-3}$ for five-step generation, $5.64\times10^{-3}$ for ten-step generation, $5.54\times10^{-3}$ for Mix-AE, and $4.31\times10^{-3}$ for Coda. Direct execution reduces error by 24.8\% relative to the five-step baseline (Fig.~\ref{fig:diagnostics}b). The extra AE update moves the result back toward baseline error, supporting the choice to execute the supervised displacement directly.

The endpoint probe in Fig.~\ref{fig:diagnostics}a complements this comparison. On \textrm{hanging\_mug}, five-step generation succeeds in 5/50 trials, Coda in 12/50, and GT-mix in 9/10. GT-mix uses a test-time expert reference and serves as a privileged diagnostic; Coda learns its correction from training demonstrations.

\subsection{Response to visual randomization}
On the 50-task Hard protocol (\textrm{demo\_randomized\_50}, minimal shader), both methods retain $K=10$. Across 2088 valid episodes per method, with 23--49 episodes per task, five-step generation and Coda reach 24.42\% and 23.86\%. The paired difference is $-0.56$ points with a 95\% interval of $[-2.71,1.64]$. The Easy-task benefit does not yield a clear gain under this visual randomization.

\section{Discussion And Limitations}
The near-equal-latency comparison shows that endpoint correction can improve closed-loop success more than an additional integration step. Results on two frozen policies support allocating inference compute between generation and correction.

The residual target combines policy error and the effects of coarse integration. Separating their contributions requires additional controls. The architecture and input comparisons also vary in capacity, training run, or checkpoint choice. In the historical $\pi_{0.5}$ continuation run, the recorded 2398 episodes give 75.62\% task-mean success, but two tasks use $K=16$ rather than the main protocol's $K=10$. This protocol difference prevents a matched comparison with Coda.

Gains vary across tasks, and the tested visual randomization yields no clear average improvement. Future work should test whether pre-execution signals can predict when correction helps and validate the measured forward-time gains in a real-robot control loop. The present evaluation is in simulation, and model-forward latency excludes communication and physical execution.

\section{Conclusion}
Coda reallocates part of a flow-matching policy's serial integration budget to one learned endpoint correction. With the VLM and action expert frozen, it improves few-step action quality and offers faster, more successful deployment points than the default ten-step policy. The near-equal-latency comparison supports this allocation of compute, and each trained corrector can serve multiple integration budgets.

\raggedbottom
\appendix\section{Implementation Details}
\label{app:details}

The appendix provides the inference procedure (Table~\ref{alg:serve}), training hyperparameters (Table~\ref{tab:hparams}), and the training and evaluation environments (Table~\ref{tab:env}).

\subsection{Inference procedure}

At test time the VLM and AE are frozen.
AE-NFE $N$ counts Euler steps.
Coda at NFE$=N$ runs the same $N$ steps as $\pi_{0.5}$-FT, then one residual forward.

\begin{table}[htbp]
\caption{Inference at AE-NFE $N$.
$\pi_{0.5}$-FT stops at $a_\theta$; Coda adds one residual; Mix-AE is ablation only.}
\label{alg:serve}
\centering
\small
\begin{tabular}{@{}p{0.95\columnwidth}@{}}
\toprule
\textbf{Input:} observation $o$, AE-NFE $N$, recipe $\in\{\pi_{0.5}\text{-FT},\mathrm{Coda},\mathrm{Mix\text{-}AE}\}$.\\
\textbf{Frozen:} $f_{\mathrm{VLM}}$, $v_\theta$.
On $\pi_{0.5}$, Coda uses residual $r_\phi$ at $t_{\mathrm{res}}=0.1$; SmolVLA does not use that time conditioning.\\
\midrule
1.\ Encode $o$ once: $z_o=f_{\mathrm{VLM}}(o)$. Cache prefix key-value tensors.\\
2.\ Draw $\varepsilon\sim\mathcal{N}(0,I)$ (use the host policy's evaluation noise stream). Set $x^{[0]}=\varepsilon$.\\
3.\ For $i=0,\ldots,N-1$:
$x^{[i+1]}=x^{[i]}-\frac{1}{N}v_\theta(x^{[i]},t_i,z_o)$, $t_i=1-i/N$.\\
4.\ Set $a_\theta=x^{[N]}$.\\
5.\ \textbf{$\pi_{0.5}$-FT:} execute $a_\theta$ and stop.\\
6.\ \textbf{Coda:} $\widehat{\Delta a}=r_\phi(z_o,a_\theta,\varepsilon,t_{\mathrm{res}})$,
execute $a_{\mathrm{exec}}=a_\theta+\widehat{\Delta a}$.\\
7.\ \textbf{Mix-AE} (ablation): form $\widetilde{x}_{0.1}=0.1\,\varepsilon+0.9(a_\theta+\widehat{\Delta a})$
with the same $\varepsilon$ as step 2, then let the frozen AE integrate from $t{=}0.1$ to $0$.\\
\midrule
Joint commands: linear interpolation, stride $16$.
$\pi_{0.5}$ eval pairs $\pi_{0.5}$-FT/Coda on the same $\varepsilon$; the stream is indexed by environment seed and replan step.
Early failure termination and evaluation video recording are disabled.\\
\bottomrule
\end{tabular}
\end{table}

\subsection{Training hyperparameters}
Table~\ref{tab:hparams} lists the training hyperparameters used in the main experiments.
We use AdamW~\cite{loshchilov2019adamw} with cosine learning-rate decay~\cite{loshchilov2017sgdr}.
The two main training stages each present 2.88M samples.
The historical continuation run matches residual training GPU-hours.

\begin{table}[htbp]
\caption{Training hyperparameters.
$\pi_{0.5}$-FT / Coda / Cont use the $50$-task RoboTwin \textrm{clean50} mixture.
SmolVLA residual uses the official $10$-task set; official SmolVLA stays frozen.}
\label{tab:hparams}
\centering
\small
\setlength{\tabcolsep}{4pt}
\begin{tabular}{lcccc}
\toprule
 & \textbf{$\boldsymbol{\pi}_{0.5}$-FT} & \textbf{Coda} & \textbf{Cont} & \textbf{SmolRes} \\
\midrule
Init & $\pi_{0.5}$ FT & FT@60k & FT@60k & official \\
Trainable & VLM$+$AE & residual only & AE & residual only \\
Global batch & $48$ & $960$ & $48$ & $256$ \\
Steps & $60$k & $3$k & $21$k & $4$ ep \\
Frames & $2.88$M & $2.88$M & $1.01$M & $4$ epochs \\
Peak lr & $2.5{\times}10^{-5}$ & $5{\times}10^{-5}$ & $2.5{\times}10^{-5}$ & $1{\times}10^{-4}$ \\
Warmup & $3$k & $150$ & $1{,}050$ & $100$ \\
Floor lr & $2.5{\times}10^{-6}$ & $5{\times}10^{-6}$ & $2.5{\times}10^{-6}$ & $2.5{\times}10^{-6}$ \\
Loss & FM (AE) & $L_1(\Delta a,a^\star{-}\mathrm{sg}(a_\theta))$ & FM (AE) & $L_1$ residual \\
$N_{\mathrm{train}}$ / $t_{\mathrm{res}}$ & -- & $5$ / $0.1$ & -- & $10$ / unused \\
Hardware & $4{\times}$A100 & $4{\times}$A100 & $4{\times}$A100 & $4{\times}$4090 \\
\bottomrule
\end{tabular}
\end{table}

For $\pi_{0.5}$, training candidates are generated by the frozen policy with $N_{\mathrm{train}}=5$ Euler steps.
On SmolVLA, AE-cont and Coda ($+\varepsilon$) are independently trained under a matched budget (per-GPU batch $64$).
The same corrector is evaluated at NFE $\in\{1,2,5,10\}$.

\subsection{Training and evaluation environments}
\begin{table}[H]
\caption{Environments for every closed-loop and latency number in the main text.
Easy and Hard use $\pi_{0.5}$; Smol Easy uses official SmolVLA.}
\label{tab:env}
\centering
\small
\setlength{\tabcolsep}{4pt}

\begin{tabular}{lccc}
\toprule
 & \textbf{Easy} & \textbf{Hard} & \textbf{Smol Easy} \\
\midrule
Suite & RoboTwin 2.0 & RoboTwin 2.0 & RoboTwin 2.0 \\
Split & \textrm{demo\_clean\_50} & \textrm{demo\_randomized\_50} & \textrm{demo\_clean\_50} \\
Tasks & $50$ & $50$ & official $10$ \\
Shader & default (rt32) & \textrm{minimal} & default \\
Target $n$ & $50$ & mid-$20$s--$49$ & $50$ \\
Seeds & $100000{+}i$ & $100000{+}i$ & $100000{+}i$ \\
$K$ & $10$ (all tasks) & $10$ (all tasks) & $50$ \\
Exec & lerp / stride $16$ & lerp / stride $16$ & $30$\,Hz $480{\times}640$ \\
FM $\varepsilon$ & det.\ (env, step) & det.\ (env, step) & i.i.d. \\
Cameras & RoboTwin default & RoboTwin default & official four \\
Metric & unweighted task mean & unweighted task mean & unweighted task mean \\
\bottomrule
\end{tabular}\\[2pt]
{\small\raggedright Default YAML pools in $100000$--$100049$ are smaller on six tasks: \textrm{adjust\_bottle} $35$, \textrm{dump\_bin\_bigbin} $32$, \textrm{pick\_diverse\_bottles} $23$, \textrm{place\_dual\_shoes} $46$, both shake $31$. Extra stable seeds fill each to $n{=}50$.
Unstable Hard resets are skipped.
On $\pi_{0.5}$, each replan samples $\varepsilon\sim\mathcal{N}(0,I)$ from seed $\mathrm{env}{\times}10007{+}\mathrm{step}$; training is unseeded. SmolVLA eval is i.i.d.
Latency: batch $1$, no compile, SDPA, warmup $10$ $+$ $40$ CUDA-sync forwards, p50 on A100 80 GB PCIe and RTX $4090$; split prefix / AE / residual.
Totals are measured independently of the stage medians.\par}
\end{table}

\end{document}